**TITLE PAGE**

# Quantitative Evidence Mining for Plausibility-Aware Biomedical AI: A Narrative Review and Conceptual Framework

Negin Sadat Babaiha[a,*], Stefan Geissler[b], Marie-Christine Simon[a], Martin Hofmann-Apitius[a,c], Marc Jacobs[a]

[a] Department of Bioinformatics, Fraunhofer Institute for Algorithms and Scientific Computing (SCAI), Schloss Birlinghoven, Sankt Augustin, Germany

[b] Kairntech SAS, 29 Chemin du Vieux Chêne, 38240 Meylan, France

[c] Bonn-Aachen International Center for Information Technology (b-it), University of Bonn, Bonn, Germany

**Correspondence to: Negin Sadat Babaiha,** Department of Bioinformatics, Fraunhofer Institute for Algorithms and Scientific Computing (SCAI), Schloss Birlinghoven, Sankt Augustin, Germany, **negin.babaiha@scai.fraunhofer.de**

## Abstract

Biomedical artificial intelligence is moving from literature retrieval toward evidence synthesis for knowledge graphs, clinical decision support, and computational models. Yet most information-extraction systems still represent findings as simple relations, discarding the quantitative and contextual detail needed for interpretation and reuse. A claim that one entity affects another is insufficient when the magnitude, unit, population, comparator, experimental conditions, uncertainty, and provenance are missing. We define quantitative evidence mining as a framework for transforming biomedical findings into structured, context-rich, and auditable evidence units. We define the core elements of an evidence unit: the claim; measured entity and property; value, unit, or scale; comparator; population; biological or clinical conditions; temporal context; uncertainty; provenance; validation results; and expert-review status. We propose an eight-stage reference architecture spanning corpus selection, entity recognition, quantity extraction, context linking, normalization, evidence-unit assembly, multidimensional plausibility assessment, and export and governance. A central principle is that plausibility should not be collapsed into a single truth label; statistical, biological, methodological, contextual, and provenance-based support should remain explicit. The framework links information extraction to evidence synthesis and computational reuse, with applications in clinical-trial analysis, biomarker research, pharmacovigilance, knowledge-graph construction, and mechanistic modelling. It is a research agenda rather than a validated end-to-end system. Progress will require annotated multimodal benchmarks, rigorous component- and workflow-level evaluation, prospective testing, transparent provenance, and sustained expert oversight.

## 1. Introduction

### 1.1 Background

Biomedical AI is shifting from information retrieval to evidence reuse: the evidence these systems extract is increasingly fed into knowledge graphs, clinical decision-support systems, and computational models, not read only by humans. This expanded role raises the standard for what such systems must capture.

### 1.2 Rationale and knowledge gap

Biomedical claims frequently depend on dosage, biomarker concentration, effect magnitude, comparator, and uncertainty. These details often appear in tables, figures, and supplementary materials rather than in narrative text. A short relation or summary may support discovery, but it is insufficient for cross-study comparison, clinical-trial prioritization, treatment decisions, or simulation parameterization when units, populations, time frames, uncertainty, and source location are absent. Biomedical AI therefore needs to move beyond relation extraction toward multimodal quantitative evidence mining. Systems should identify values and units and link them to the measured entity, property, population, intervention, comparator, time point, uncertainty, and exact source evidence. Several communities already standardize parts of this problem — PICO/EBM extractors, measurement shared tasks, and domain catalogues such as GWAS, PGS, and Open Targets. What remains missing is a representation that assembles these elements across text, tables, and figures and pairs them with explicit, multidimensional plausibility assessment and process provenance. That assembly, rather than any single field, is the gap we address. We use the term quantitative evidence mining to denote this assembly-and-plausibility problem as framed in this paper, rather than to name an already-established field.

### 1.3 Objective

This narrative review introduces the biomedical quantitative evidence unit and a plausibility-aware reference architecture for deriving such units from text, tables, and figures. It synthesizes relevant methods, defines the information required for auditable reuse, identifies biomedical applications, proposes evaluation levels, and outlines a staged research agenda. Automated plausibility assessment is positioned as decision support rather than a replacement for expert judgment.

We present this article in accordance with the Narrative Review reporting checklist.

## 2. Methods

### 2.1 Literature search and source selection

This review used a narrative approach to integrate methodological, conceptual, and applied literature relevant to quantitative biomedical evidence mining. Literature identification was conducted iteratively between July and September 2026 using Google Scholar, backward citation searching from relevant publications, and targeted searches of authoritative regulatory, standards, and biomedical-resource websites. An additional structured PubMed search was conducted on 18 September 2026, covering the database from inception to the search date without restrictions on publication date, language, article type, species, or full-text availability.

The PubMed search used five complementary concept groups covering: (1) foundational biomedical information and relation extraction; (2) quantitative and measurement extraction; (3) large-language-model and natural-language-processing approaches to extracting clinical-trial and review data; (4) extraction from tables, figures, multimodal documents, and full-text articles; and (5) provenance and auditable evidence representation. This was expert-guided narrative synthesis, not a systematic map: sources were identified and selected by a single reviewer (NSB), without dual screening, a PRISMA flow, or risk-of-bias assessment. The PubMed search served to verify that the concept groups recovered an a priori seed set of known foundational papers, and we do not claim exhaustive coverage. PubMed-indexed foundational papers already cited in the manuscript were used as an a priori seed set to assess the sensitivity of the search concepts.

Sources were considered for inclusion when they provided a relevant methodological contribution, empirical evaluation, benchmark, authoritative standard, biomedical application, database or resource description, or conceptual foundation for quantitative evidence extraction, representation, validation, or reuse. Foundational work from general scientific information extraction was retained when its methods were directly transferable to biomedical evidence mining. Sources without a direct methodological, conceptual, or applied connection to these topics were excluded.

Potentially relevant sources were assessed by NSB based on their titles and abstracts and, where necessary, their full texts. Reference lists of relevant publications were examined to identify additional sources. The selected literature was organized around biomedical relation extraction, measurement and context extraction, large-language-model and multimodal information extraction, provenance and evidence representation, plausibility assessment, evidence-aware knowledge graphs, and downstream biomedical applications.

Because this was a narrative review, no independent duplicate screening, formal risk-of-bias assessment, or quantitative meta-analysis was performed. The review does not claim exhaustive coverage of all publications within the included methodological fields.

Table 1. Search strategy summary

| Items | Specification |
|---|---|
| Date of search | Iterative searches conducted between July and September 2026; final PubMed coverage-verification search conducted on 18 September 2026 |
| Databases and other sources searched | PubMed; Google Scholar; backward citation searching from relevant publications; authoritative regulatory, standards, and biomedical-resource websites |
| Search terms used | Concept groups covering biomedical information and relation extraction; quantitative and measurement extraction; LLM/NLP-based extraction of trial and review data; extraction from tables, figures, and multimodal documents; provenance and auditable evidence representation. The complete PubMed strategy is provided in Supplementary Appendix 2. |
| Time frame | PubMed searched from database inception to 18 September 2026; no predefined publication-date restriction was applied to the narrative literature search |
| Inclusion and exclusion criteria | Methodological studies, empirical evaluations, benchmarks, reviews, conference papers, standards, database or resource papers, and biomedical applications directly relevant to quantitative evidence extraction, representation, validation, or reuse were eligible. Duplicate sources and publications without a direct methodological, conceptual, or applied connection were excluded. |
| Selection process | Candidate sources were assessed by NSB based on titles and abstracts and, where necessary, full texts. Additional sources were identified through backward citation searching. No independent duplicate screening was undertaken. |
| Additional considerations | Topic-driven narrative synthesis; foundational non-biomedical methods included when directly transferable; no meta-analysis or formal risk-of-bias assessment |

### 2.2 Scope and narrative synthesis

The literature is organized thematically around semantic relation extraction, measurement and context extraction, multimodal evidence recovery, evidence-unit representation, plausibility assessment, biomedical use cases, and evaluation. The synthesis is conceptual and explanatory; it does not estimate pooled effects or claim exhaustive coverage. Evidence from primary studies, methodological papers, standards, databases, regulatory sources, and selected reviews is used to motivate requirements and illustrate gaps.

### 2.3 Use of AI-assisted tools in manuscript preparation

SciSpace and the research or deep-search functions available in ChatGPT (OpenAI) and Claude (Anthropic) were used between July and September 2026 to support literature discovery, preliminary source summarization, and organization of the literature; specific model versions were not systematically recorded. ChatGPT and Claude were additionally used for structural refinement, condensation, and language editing of author-written text. Grammarly and QuillBot were used for grammar, spelling, style checking, and limited paraphrasing.

All references suggested by these tools were manually assessed for relevance and checked against the original publications. The authors manually verified all citations, bibliographic information, scientific claims, reported data, and quantitative values. AI-generated summaries were not used as primary evidence, and all final decisions concerning source selection, interpretation, and manuscript content were made by the authors, who take full responsibility for the manuscript.

## 3. From biomedical text mining to evidence representation

### 3.1 Semantic predications and scalable relation extraction

Early efforts in biomedical text mining focused on distilling semantic predications – assertions about entities related by specific functions or interactions. SemRep (2) grounded text in semantic categories and produced normalized triples such as Drug-TREATS-Disease. SemMedDB (3) scaled this approach to PubMed by operationalizing semantic predications from the UMLS, demonstrating that much of the literature could be distilled to computable triples. In practice, this entailed extracting triples that were relevant to a particular application such as drug repurposing for COVID-19 and subsequently ranking and filtering them using KG completion methods (4), (5). However, such triples typically encode very limited information about the quantity, qualification, and context of the interaction, which are essential for downstream analyses such as clinical trial enrichment (48), (68).

Machine learning approaches to relation extraction provided a more scalable alternative to rule-based semantic indexing, but require extensive annotated training corpora (8), (9). Our own relation extraction model achieved up to 67% F-scores on relations with sufficient training data but dropped to around 50% on low-frequency relations (8). Overall, these methods provide the critical entity-relation-entity backbone for subsequent quantitative evidence extraction but were not designed to represent quantitative information and its provenance. Figure 1 provides an overview of the evolution of biomedical text-mining methods.

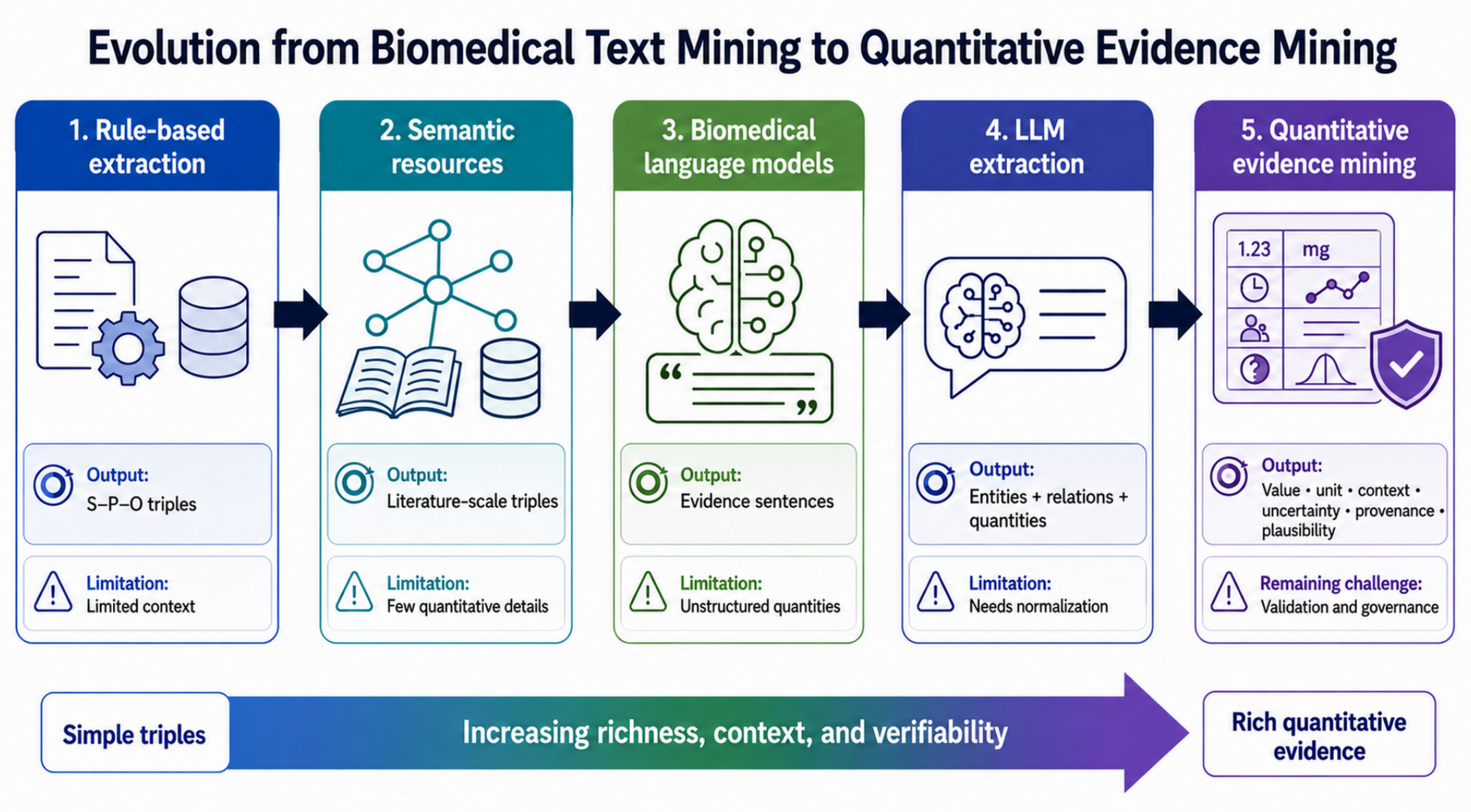

**Figure 1. Conceptual evolution from biomedical text mining to quantitative evidence mining.** The stages are not necessarily mutually exclusive – rule-based methods, semantic resources, biomedical language models, and large language models can be combined in end-to-end pipelines. Quantitative evidence mining encompasses all previous approaches while additionally retaining quantitative values, units, contextual qualifiers, evidence for plausibility, uncertainty, and provenance information.

### 3.2 Biomedical language models and evidence extraction

Zhao et al. articulated biomedical evidence engineering as a process in which data-driven insights are evaluated against the published literature to determine their textual grounding (12). Their framework exploits BERT-style evidence-sentence retrieval to identify relevant text snippets for each insight (12), (13), (14). Once again, this creates an evidential basis for a claim but does not encode much information about the quantities, contexts, and provenance that must be retained for downstream evidence synthesis, KG construction, or clinical utility.

A similar observation can be made about the clinical information extraction literature, in which claims are often represented as pairs of entities related by a biomedical interaction (17). The implicit assumption is that the textual context encodes all necessary information about the quantities, uncertainties, and qualifications of the interaction. However, this is rarely the case: a claim about a vaccine's association with protection against COVID-19 encodes much less information than the full set of supporting evidence typically reported in a clinical trial. Specifically, the trial would report the vaccine regimen (such as BNT162b2), the population (16+ years), the study endpoint (confirmed symptomatic COVID-19), the effect size (efficacy), the comparator (placebo), the confidence interval, and the statistical significance (p-value). The full claim would thus be 'Two doses of BNT162b2 vaccine 21 days apart confer high efficacy against confirmed symptomatic COVID-19 in participants 16 years of age and older' (15), (16).

LLMs offer additional capabilities, as they can encode the qualitative and quantitative context of a claim while retaining an awareness of alternative formulations and linguistic variations. Specifically, LLMs can recognize negations, uncertainties, comparisons, and modifiers such as dose schedules or time points. GPT-families and other LLMs have demonstrated proficiency in biomedical information extraction including span detection, entity recognition, and relation extraction with few examples (19)-(21), (42).

However, this proficiency is not always reflective of their fitness-for-purpose: models often fail to capture critical quantitative context or erroneously apply a quantity to the wrong entity (10), (18), (22), (23), (45), which can compromise downstream analyses if such information is not available elsewhere. In particular, an assertion about an intervention's efficacy may encode much less information than is required for evidence synthesis, clinical decision-support, or simulation modeling. Specifically, such a claim typically fails to encode who the intervention affects (which population), what it affects (which endpoint), by how much, under what circumstances, and with what degree of confidence.

## 4. Measurement and quantitative information extraction

Having established why evidence extraction goes beyond relations and sentences, we turn to the core technical issue of quantitative evidence mining: how to detect and represent measurements and their properties to enable subsequent evidence synthesis, KG construction, and downstream analyses. Specifically, we are interested in capturing what has been measured, how much it has changed, in which units and populations, and with what degree of confidence.

### 4.1 Measurement extraction as a scientific information extraction task

Scientific documents contain a wealth of quantitative information, but its representation in text is highly non-uniform. At a minimum, any measurement encodes a value and a unit but may also describe the measured entity, the measured property, and various qualifiers that affect the interpretation of the measurement. This task has been recognized in the MeasEval shared task, which targeted the extraction of counts, measurements, and related contextual information as part of the SemEval-2021 Task 8 (24).

MeasEval included a range of subtasks, including the detection of units, value-modifiers, measured entities, properties, qualifiers, and quantity-context relations.

The need to track quantity and context is particularly pertinent in biomarker research, in which the choice of biological fluid, assay, population, disease state, thresholds, and timing can drastically modulate a measurement (25-29).

### 4.2 Rule-based, hybrid, and machine-learning systems for quantity extraction

Quantity extraction has long been a staple of information extraction with rule-based, hybrid, and machine-learning methods. Contextual linguistic patterns can be leveraged to identify mentions of percentages, ranges, P values, odds, and hazard ratios with high precision, and learned models can associate such quantities with relevant entities and properties (37-43). Hybrid methods use rule-based methods for candidate detection and learned entity-relation models (37), (45).
However, the document layout can create challenges for quantity extraction, notably by separating the quantity from its context.
For example, a dose mentioned in the Methods section may be separated from the relevant outcome mentioned in the Results section, leading to spurious associations if the two are erroneously linked.

### 4.3 LLM-based quantitative information extraction

Although LLMs can associate numbers with contexts that differ in wording and extract records from sentences, tables, or figures (21,44), fluent outputs do not reliably attach numerical values correctly: the model might have attached a value to the wrong endpoint, have mixed up the baseline and the follow-up, or generalized a subgroup analysis to the whole cohort (10), (18), (22), (23), (45). Extraction using an LLM should occur in a controlled workflow that preserves source spans, prompts, model versions, segmentation decisions, decoding settings such as temperature, and output schemas.

Recent systems take inspiration from this direction, identifying and normalizing quantities and using question-answering to recover measurement contexts (44), or combining rule-based candidate detection with LLM extraction and retrieval-based validation (45). These techniques help move from quantity detection to source-grounded, context-preserving evidence units but do not yet prove end-to-end reliability. Table 2 maps what established systems already capture against what they omit. Parts of the representation — value, unit, population — are standardized in adjacent communities; the contribution of the quantitative evidence unit is their assembly with explicit multidimensional plausibility and process provenance, not the invention of any single field. Recent work has begun to operationalize parts of this direction. Mortezaagha et al. (65) developed a schema-constrained system for extracting methodological, laboratory, and outcome variables from full-text biomedical PDFs. The system combined typed schemas, controlled vocabularies, caption-aware document segmentation, deterministic record assembly, and sentence-level evidence capture to produce auditable structured records. This work demonstrates the feasibility of provenance-rich extraction from heterogeneous full-text documents, although broader evaluation of field-level accuracy, cross-modal quantity–context linking, and multidimensional plausibility assessment remains necessary.

Table 2. Related work

| System / standard | Layer | What it captures | What it drops (that the evidence unit adds) |
|---|---|---|---|
| SemRep / SemMedDB | Relation IE | Normalized entity–relation–entity triples | Quantity, units, population, comparator, uncertainty, provenance |
| MeasEval, GROBID-quantities, CQE, Quinex | Measurement IE | Value, unit, measured entity/property, local qualifiers | Binding to a biomedical claim, multi-axis plausibility, process provenance |
| EBM-NLP, Evidence Inference, Trialstreamer, RobotReviewer, TrialSieve | RCT / PICO extraction | PICO elements, some effect directions | Harmonized value+unit, cross-source consistency, audit trail |
| SciFact, HealthVer, COVID-Fact | Claim verification | Supported / refuted label vs. evidence | Structured quantity, context fields, non-collapsed quality dimensions |
| Nanopublications, micropublications, BEL | Structured claims | Assertion + provenance + attribution | Quantitative value/unit semantics, statistical-coherence checks |
| Open Targets evidence model | Target–disease evidence | Typed evidence, scores, source | Full measurement context + explicit multidimensional plausibility |

| System / standard | Layer | What it captures | What it drops (that the evidence unit adds) |
|---|---|---|---|
| GWAS Catalog / PGS Catalog | Domain schemas | Variant/score fields, effect sizes, ancestry | Cross-domain reuse, literature-derived units, flag-don't-discard checking |
| ChEMBL activities, OMOP, CDISC-SDTM | Structured records | Standardized measurements within one silo | A representation spanning text, tables and figures with provenance of the extraction |
| Biomedical quantitative evidence unit (this paper) | Representation | Value+unit + entity/property + comparator + population + condition + time + uncertainty + provenance + per-dimension plausibility + review status | — (the assembly across silos is the contribution) |
| Schema-constrained full-text PDF extraction | Full-text evidence extraction | Structured study variables, controlled schemas, sentence-level evidence, caption-aware processing, provenance | Claim-centred assembly across modalities, explicit missingness, comparator and uncertainty linkage, and separate multidimensional plausibility assessments |

## 5. Biomedical quantitative evidence units

Section 4 showed that extracting numbers is not enough. To become usable evidence, a quantity must be connected to what it measures, where it was reported, the conditions under which it holds, and the uncertainty reported with it. We therefore define the biomedical quantitative evidence unit as the core representation: a structured container linking a claim to its value, unit, measured entity and property, context, comparator, uncertainty, provenance, validation results, and expert-review status.

### 5.1 Components of a biomedical quantitative evidence unit

A biomedical quantitative evidence unit is a structured representation that connects a scientific claim to the quantitative evidence supporting it. At a minimum, it should preserve the claim structure: subject or intervention, predicate or direction of effect, object or outcome, together with the entity measured, property measured, value and unit or scale, comparator, population, biological or clinical condition, temporal context, uncertainty, provenance, and exact source evidence span. Extraction confidence, multidimensional validation results, and expert-review status should be represented separately rather than collapsed into a single plausibility label (Table 3). We define one evidence unit as a single quantitative estimate for one measured property in one population or arm, under one comparator and one time point; a source reporting several endpoints, arms, or time points therefore yields several evidence units. Each field should also carry a production status — verbatim, normalized, calculated, model-inferred, or human-added — because a normalized unit or a derived effect estimate is not equivalent to text copied directly from the source.

Table 3. Core components of a biomedical quantitative evidence unit

| Component | Field | Illustrative example |
|---|---|---|
| Claim | Claim identifier | QEU-001 |
| Claim | Claim structure | Drug X — REDUCES — serum IL-6 concentration relative to placebo |
| Measurement | Measured entity | Interleukin-6 (IL-6) |
| Measurement | Measured property | Serum concentration |
| Measurement | Value and unit/scale | 35% relative reduction |
| Context | Intervention or exposure | Drug X |
| Context | Comparator/reference | Placebo |
| Context | Population | Older adults; sex and gender recorded as distinct variables when reported |
| Context | Biological material | Serum |
| Context | Experimental or clinical condition | Disease condition and assay not reported |
| Context | Temporal context | After 12 weeks |
| Uncertainty | Uncertainty measure | 95% confidence interval: 20%–47% |
| Provenance | Source document and location | Illustrative Study X, source Figure 2 |
| Provenance | Source evidence span | Exact sentence, table cell, or figure region |

| Component | Field | Illustrative example |
|---|---|---|
| Extraction | Extraction confidence | To be reported separately for extracted fields and contextual links |
| Assessment | Multidimensional assessment | Source grounding: confirmed; unit/scale consistency: valid; contextual completeness: incomplete; statistical coherence: not assessed; biological plausibility: no conflict detected; cross-source consistency: not assessed |
| Governance | Expert-review status | Not reviewed |

All values and source labels in Table 3 are examples. Assessment dimensions should be kept separate as source grounding, extraction correctness, contextual completeness, statistical coherence, biological plausibility, cross-source consistency are related but different properties; we deliberately do not collapse them into a single overall evidence-quality score.

These assessments are not assumed to be extracted directly from the source. They are generated through subsequent validation steps as described in Section 6.7. Similarly, not all evidence-unit fields will necessarily be available in a given source. Missing comparators, uncertainty measures, or provenance information should be represented explicitly as “not reported in the source”, rather than implicitly or omitted, or a basis for discarding the claim. Distinct missingness states should be preserved — “not reported in the source”, “not applicable”, “not extracted”, “extraction failed”, and “not yet reviewed” — because they carry different scientific and workflow meanings and should not be collapsed.

The evidence unit thus extends measurement extraction by tying a value and unit to the biomedical claim it supports, while preserving the information necessary to interpret and audit that claim.

Applied to the BNT162b2 example, the evidence unit ties the reported efficacy estimate to the intervention, comparator, population, endpoint, time point, uncertainty interval, underlying event counts, and precise source location. Plausibility assessment then determines if these are correctly linked, internally coherent, and supported by the cited evidence. The full representation is available in Supplementary Appendix 1.

**Worked example — BNT162b2 primary efficacy (Polack et al., NEJM 2020):**

| Field | Value |
|---|---|
| Claim | Two 30-µg doses of BNT162b2, 21 days apart, are efficacious against confirmed symptomatic COVID-19 |
| Measured entity / property | BNT162b2 vaccine / vaccine efficacy (relative risk reduction) |
| Value and scale | 95.0% efficacy (relative, %); field status: verbatim |
| Uncertainty | 95% CI 90.3–97.6 |
| Comparator | Placebo |
| Population | Participants ≥16 years, no prior evidence of infection |
| Endpoint | Confirmed symptomatic COVID-19 ≥7 days after dose 2 |
| Underlying counts | 8 vs. 162 cases (vaccine vs. placebo) |
| Temporal context | ≥7 days post dose 2; median follow-up 2 months |
| Provenance | NEJM 2020;383:2603–2615, Results / efficacy table; sentence + table cell |
| Plausibility (per dimension) | source-grounded: pass; statistically coherent: pass (counts ↔ efficacy ↔ CI); completeness: complete; biological plausibility: no-reference; review status: expert-confirmed |

Represented this way the estimate is comparable (efficacy, endpoint, population are explicit), traceable (span + table cell), and checkable (event counts reconcile with the interval) — none of which a Drug–TREATS–Disease triple or a single efficacy sentence preserves.

## 5.2 Non-exclusive facets of biomedical quantitative evidence

Biomedical quantities can be characterized from a number of perspectives. The same observation could be of particular value in a study, be related to particular domains in biomedicine, and potentially have been measured in different quantitative forms. Figure 2 summarizes these non-exclusive facets, while Supplementary Appendix 1 provides a more detailed organization across the three axes of evidence role, biomedical domain and quantitative form.

These facets were defined to support annotation across dimensions, rather than assignment to a specific quantity type. They represent requirements depending on the context, rather than information that will always be present. For fields that are not present, these should be recorded explicitly as “not reported in the source” rather than being omitted or estimated.

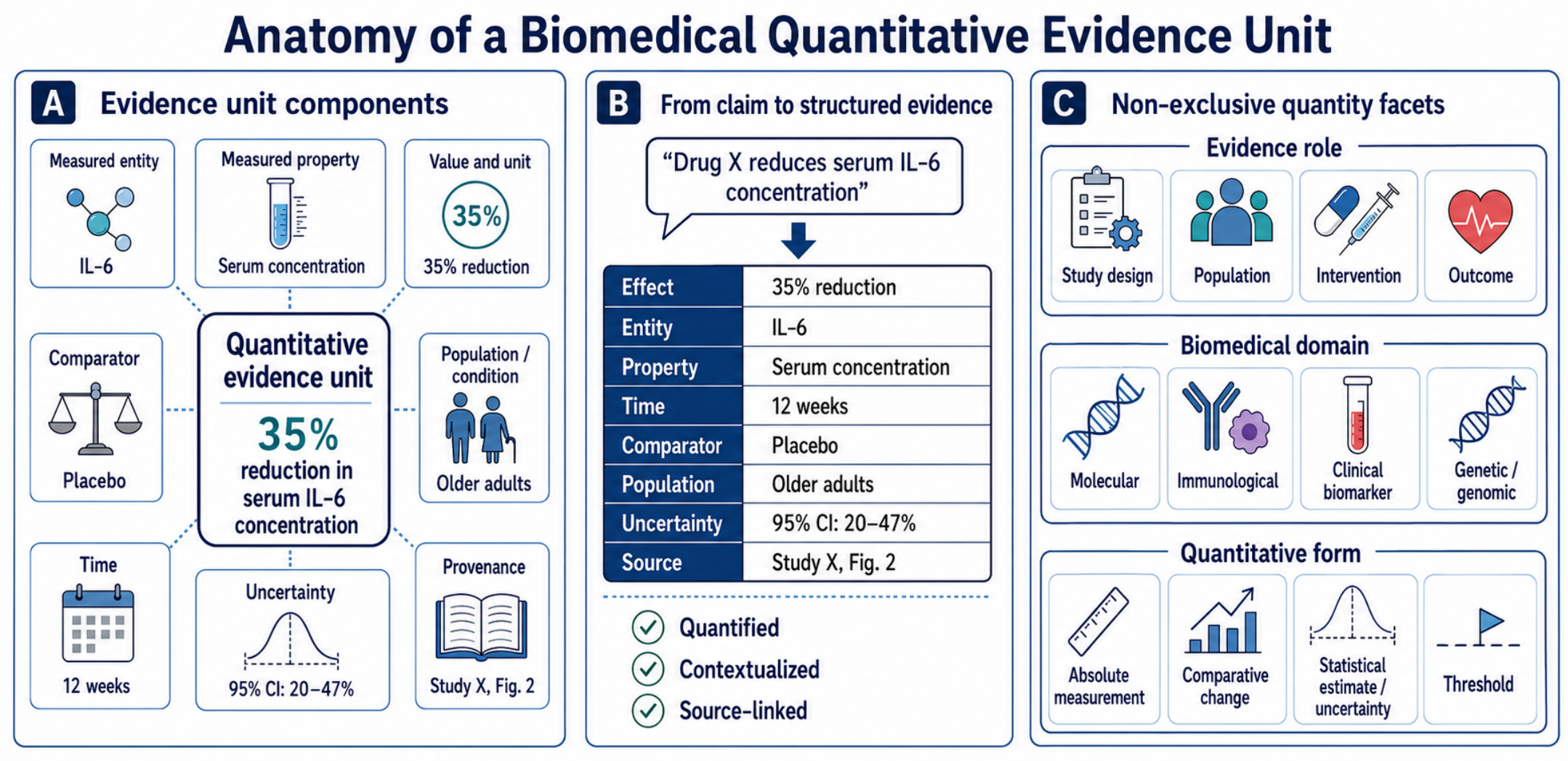

**Figure 2. Anatomy of a biomedical quantitative evidence unit.** (A) Core components connect the measured entity and property with the value, comparator, population or condition, time, uncertainty, and provenance. (B) A narrative claim is converted into a quantified, contextualized, and source-linked structured record. (C) Quantities are described through complementary and non-exclusive facets of evidence role, biomedical domain, and quantitative form. All values and source labels shown are illustrative.

## 6. Plausibility-aware framework

Having defined the evidence unit, the next question is how such units are produced and checked. We propose a layered reference architecture (Figure 3) that turns raw sources into structured, traceable, plausibility-checked units. It is not a fixed pipeline: different implementations may use rule-based, machine-learning, LLM, or hybrid methods at each layer.

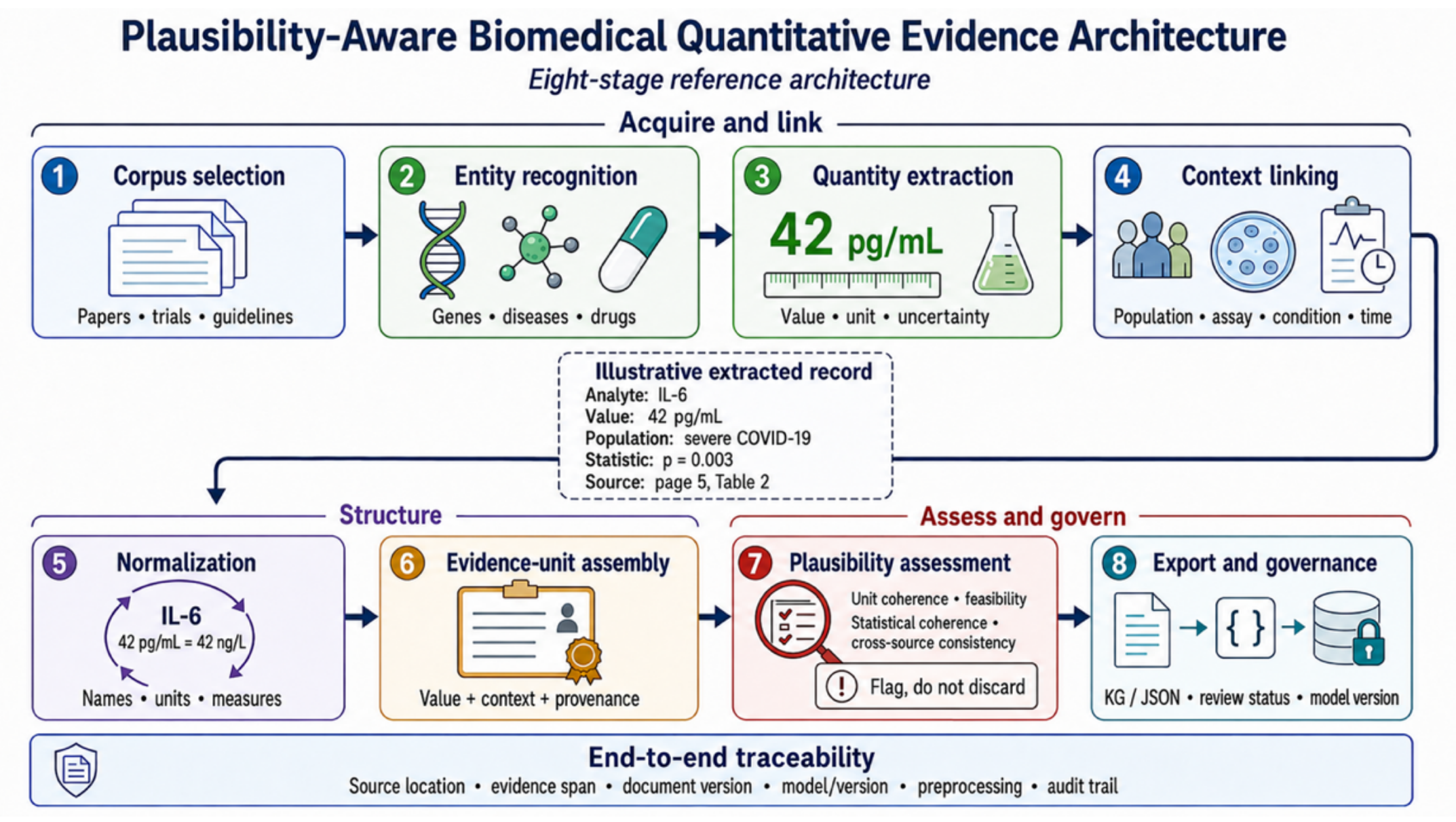

**Figure 3. Eight-stage reference architecture for plausibility-aware biomedical quantitative evidence mining.** The workflow proceeds from corpus selection, entity recognition, quantity extraction, and context linking to normalization, evidence-unit assembly, multidimensional plausibility assessment, and export and governance. JavaScript Object Notation (JSON) is shown as one structured export format. End-to-end traceability preserves source location, evidence span, document version, model and software version, preprocessing, and the audit trail. Questionable evidence is flagged for review rather than automatically discarded.

### 6.1 Corpus generation and source selection

The first step in the protocol includes the definition of the source corpus. This corpus can consist of various types of biomedical data depending on the question. It can include PubMed abstracts, PMC full-text documents, clinical trials, systematic reviews, preprints, guidelines, regulatory documents, or additional files. The choice of the corpus depends on the target evidence question, for example, vaccine response, ageing biomarkers, immunology-related disease, post-viral condition, drug efficacy, or neurodegenerative comorbidity.

### 6.2 Biomedical entity and concept recognition

The second level detects biomedical entities in the texts and maps them to controlled vocabularies. The entities include genes, proteins, diseases, chemicals, drugs, phenotypes, biomarkers, pathways, variants, cell types, tissues, organisms, interventions, and outcomes. The resources include PubTator (6), (7), UMLS, MeSH, MONDO, HPO, ChEBI, DrugBank, and Gene Ontology.

### 6.3 Quantity and unit extraction

The third layer is the extraction of numeric values and units including exact values, ranges, percentages, fold changes, p-values, confidence intervals, odds ratios, hazard ratios, risk ratios, sample sizes, durations, concentrations, doses, titers, thresholds, etc. This layer can be built on top of MeasEval (24), GROBID-quantities (41), CQE (37) and Quinex (44). Importantly, quantity extraction should not only be limited to narrative text but also expand to the visual modalities, as important clinical and biomarker trends are often locked in figures, forest plots and survival curves. Conventional optical character recognition (OCR) engines such as Tesseract (64) are good for digitizing text blocks and simple tables. However, for extracting complex visual data like data points from a scatter plot or hazard ratios from a forest plot, the framework leverages Vision-Language Models (VLMs) such as BiomedGPT (46). These multimodal models can understand visual layouts, read axis scales, and directly map plotted symbols to structured quantities. For visual extraction, the provenance metadata of the produced evidence unit explicitly notes the modality (e.g. 'figure panel', 'visual plot', or 'caption'), for a full audit trail. Quantitative recovery from figures can also be approached through specialized computer-vision pipelines. Cardaras et al. (66) combined figure classification, optical character recognition, chart-component detection, and axis-based interpolation to reconstruct numerical values from biomedical bar charts. In their proof-of-concept evaluation, 81.2% of reconstructed bar values were within 5% of the manually annotated reference values. The remaining errors, particularly those caused by OCR digit confusion and ambiguous bar boundaries, illustrate why visual quantities should retain region-level provenance and extraction confidence rather than being treated as exact observations by default.

### 6.4 Context extraction

The fourth layer links each quantity back to the context in which it was measured: what entities and properties were measured (populations, species, tissues, cell lines, assays), the disease stage, treatment arm, comparator, time point, and endpoint. Without this context a value is potentially uninformative or worse, misleading. For genetic evidence the context should also include genome build, effect allele, definition of phenotype, ancestry, sex distribution, methods used for genotyping or imputation, statistical model and covariates. Context extraction is itself an active research problem rather than a simple post-processing step. Tang et al. (67) developed CELESTA, a semi-supervised multi-task framework for identifying contextual categories such as cell type, cell line, disease, and intracellular location in biomedical text. Their results demonstrate that contextual qualifiers can be modelled explicitly, including cases outside known categories. For quantitative evidence mining, these contextual predictions must subsequently be linked to the correct quantity, entity, property, and claim, with uncertainty retained for each attachment.

### 6.5 Normalization and harmonization

The fifth level ensures normalization of units of measurement, quantity, entity names, and statistical measures: for example, reconciles IL-6 and interleukin-6, pg/mL and ng/L, odds, hazard, and risk ratios, and fold changes versus absolute concentrations. It also standardizes variant identifiers and genome builds in genetics while preserving the orientation of effect alleles, thus preventing incorrect pooling of estimates.

### 6.6 Evidence unit construction

The sixth level bundles these components into a structured whole that can be used in dashboards, systematic reviews, evidence tables, retrieval-augmented generation systems, and decision support. At this stage, the record can be inspected and compared, but it is not yet fit for use until plausibility and provenance checks have been applied.

### 6.7 Plausibility checking

The seventh layer conducts a multi-faceted assessment of plausibility rather than a single true/false label across different dimensions, including source grounding, extraction confidence, unit and scale consistency, contextual completeness, assay compatibility, statistical coherence, biological plausibility, cross-source consistency, and outlier status, reporting distinct results and review flags per dimension rather than one aggregated truth value.

When applied to the IL-6 unit in Figure 2, this would involve a source grounding check whether the reported 35% reduction indeed appears in the identified passage or figure, a contextual check comparing it only to compatible populations, assays, materials, and time points, and a missing context check that would note the absence of information without discarding the unit.

Crucially, plausibility should not be conflated with scientific truth: a value may be correctly extracted, traceable, consistent internally yet originate from a biased or underpowered study (1), and an apparently outlying value may reflect a discovery rather than an extraction error. The automatic assessment of plausibility should therefore be used as a decision support tool while retaining uncertainty and expert judgment. Each dimension has its own input, decision procedure, output type, and storage (Table 4). Two consequences follow: a value with no established reference range returns ‘no-reference’ rather than ‘implausible’, and two disagreeing values are marked ‘incomparable’ rather than ‘discrepant’ until an assay/material/unit commensurability check passes. For each statistical check, the applicable effect measures, eligible input type, tolerance, and not-applicable condition are specified in Supplementary Appendix 1.

Table 4. Dimensions of plausibility assessment for biomedical quantitative evidence units

| **Dimension** | **What is inspected** | **Check / decision procedure** | **Output** | **How the flag is stored** |
|---|---|---|---|---|
| Source grounding | Extracted value vs. cited source span or figure | Verify the original region or cell and record any unit conversion or derivation; a string or number match alone is insufficient for OCR-derived, converted, or reconstructed values | pass / fail / unknown | Per-unit boolean + span offset |
| Extraction correctness | Value–entity–property attachment | Rule + human spot-check | pass / fail / unknown | Per-field confidence score |
| Contextual completeness | Presence of population, comparator, time, assay | Required-field checklist | complete / partial (list missing) | “Not reported in source” token per field |
| Statistical coherence | CI vs. point estimate; $z = \beta/SE$ vs. p; n vs. precision | Deterministic algebraic checks | pass / fail / not applicable (no dispersion reported) | Numeric residual + flag |
| Biological plausibility | Value vs. reference range where one exists | Range check; else “no prior available” (never an LLM oracle) | in-range / out-of-range / no-reference | Flag + which prior was used |
| Cross-source consistency | This value vs. other units for the same entity | Compare only after a commensurability check (same assay / material / unit) | consistent / discrepant / incomparable | Pairwise flag + commensurability basis |

### 6.8 Provenance, modality, and process traceability

In addition to the content, it is important to retain the production history of the unit (source document, section, sentence, table, figure, caption, extraction model, prompt, decoding settings such as temperature, model version, execution date, segmentation, and post-processing) since the same document may generate different results depending on the settings and versions of the model. The PROV-O standard (47) can be used to represent the provenance of generating entities, activities, and agents, supporting audit trails and reproducibility; the provenance should also encode the unit's modality (i.e., sentence, table cell, figure panel, caption, or supplement). Provenance coordinates should be modality-specific: section, paragraph or sentence, and character span for prose; table, row, column, and header path for tables; and figure, panel, caption, and image-region coordinates for figures. Source version — document version and correction or retraction status — should be recorded separately from extraction version — extraction-model version, schema version, and validation-rule version — because these change independently and are both required for reproducible reprocessing.

## 7. Biomedical applications and downstream uses of quantitative evidence mining

Quantitative evidence mining is relevant wherever quantitative biomedical claims must be compared, validated, or reused across heterogeneous documents, populations, platforms, and biological scales—from evidence synthesis in trials and biomarker research to evidence-aware KGs to simulation-ready parameters. The same evidence-unit structure helps each.

### 7.1 Clinical trial analysis

In clinical-trial analysis, evidence units capture the elements needed to compare studies—eligibility criteria, intervention dose and schedule, treatment duration, comparator, endpoints, effect estimates, confidence intervals, adverse events, and follow-up (35), (48). Trial KGs organize studies, interventions, and outcomes; evidence units add the numerical evidence attached to design and results, distinguishing that an intervention was tested from a reusable statement of what was tested, in whom, and with what result.

TrialSieve (68) provides a recent example of extending conventional PICO extraction toward treatment-group-based representations for meta-analysis and drug repurposing. Its annotation framework captures a broader set of trial entities and organizes them hierarchically to support quantitative comparisons. Such systems provide an important foundation, while quantitative evidence units additionally require explicit value–endpoint–group attachment, units or scales, uncertainty, precise source location, and separate validation results.

### 7.2 Biomarker evidence mining

Biomarker evidence mining targets measurable indicators—cytokine concentrations, antibody titres, omics values, diagnostic thresholds, and subgroup effects (49)—whose meaning depends on biological material, assay, disease stage, subgroup, comparator, time point, and uncertainty. "IL-6 is associated with disease severity" is far less informative than a unit preserving the IL-6 concentration, unit, sample type, group, comparator, endpoint, and confidence interval, which lets plausibility checks catch unit mismatches or out-of-range values.

### 7.3 Complex disease applications and quantitative evidence-aware KGs

Complex-disease and comorbidity research is a natural application, because an association is rarely interpretable without magnitude and context. In COVID-19 comorbidity research, neurological or neurodegenerative risk estimates depend on cohort size, severity, follow-up, age, vaccination status, outcome definition, comparator, and effect size—preserving these allows comparison across heterogeneous studies rather than reduction to a generic disease–disease link (30)-(34).

These requirements suggest a shift from conventional KGs, which represent entities and relations, to evidence-aware KGs that connect a claim to its value, unit, measured property, context, uncertainty, provenance, validation dimensions, and review status (50). The COMMUTE project (https://www.commute-project.eu/en/about.html), investigating links between SARS-CoV-2 infection and neurodegenerative comorbidities, illustrates this: related work integrated curated databases with text-mining KGs, built a manually curated causal network (9), and showed multimodal LLMs can extract mechanistic triples from figures (11)—resources that support navigation but not systematically preserve quantitative context for meta-analysis or calibration.

The protein-kinase-inhibitor KG of Jackson et al. (36) shows this transition externally: it integrates ChEMBL, PubMed, ClinicalTrials.gov, and FAERS, weights drug–condition edges by supporting-literature characteristics, and attaches measures such as hazard ratio, progression-free survival, and overall survival. This demonstrates evidence weighting while underscoring the need to normalize values, preserve uncertainty and context, and assess plausibility—which evidence units provide.

### 7.4 Pharmacovigilance and regulatory assessment

Pharmacovigilance and regulatory assessment depends on quantitative evidence for safety and benefit–risk decisions—dose, exposure, adverse-event frequency and seriousness, time-to-onset, thresholds, efficacy outcomes, risk estimates, confidence intervals, and reporting context (51), (52). A bare drug–event relation is insufficient; magnitude, denominator, uncertainty, source, and limitations must remain visible, and provenance-linked units can support comparison and transparent expert review without replacing regulatory judgment.

### 7.5 From quantitative evidence units to simulation-ready parameters

Quantitative evidence mining can also support computational modelling, particularly agent-based modelling, by turning reported measurements into candidate model parameters. Agent-based models represent systems through interacting components—cells, pathogens, immune mediators, or patients—and are useful when system-level behaviour emerges from many local interactions. A quantitative evidence unit makes a candidate parameter traceable—recording what was measured, in which unit and material, at which time point, in which population, and with what uncertainty—so values can be mapped to model variables, harmonized, represented as ranges, and labelled for calibration or validation. Immune-system simulators such as UISS illustrate this need (53), (54), (55), (56); evidence mining does not replace mechanistic modelling but makes the link between literature and modelling assumptions transparent. A recent quantitative systems pharmacology application illustrates how structured extraction can support model calibration while retaining human oversight. Eliason and Popel introduced MAPLE (69), in which schema-constrained records separate literature-derived values from modelling decisions, connect extracted values to source evidence, and apply automated validators for source matching, citation resolution, and executable checks. Its application to one disease-specific model demonstrates the value of provenance-rich parameterization, while its limited evaluation also reinforces the need for cross-domain benchmarks and prospective validation.

### 7.6 Sex- and gender-aware evidence mining

Sex and gender are best treated as a cross-cutting lens: they should be represented as distinct population variables when reported, because reference ranges, pharmacokinetics, treatment effects, and adverse-event profiles differ across populations. A meta-analysis of 86 FDA-approved drugs found women on the same dose often had higher blood concentrations, slower elimination, and more adverse reactions (57), and evidence gaps persist in cardiovascular care (58). A pipeline that preserves population composition can mark a claim derived mainly from one population as under-validated for another rather than silently generalizing it.

### 7.7 Quantitative genetic evidence mining

Quantitative genetics is especially suitable, since a variant–trait relation is uninterpretable without statistical and population context. A reusable unit should preserve the variant identifier and genome build, effect and non-effect alleles, allele frequency, phenotype definition, effect estimate, standard error or confidence interval, p-value, sample size, ancestry, sex composition, statistical model, covariates, and replication status; for polygenic scores, the variants and weights, development and validation populations, and predictive-performance measures (59), (60), (61), (62), (63).

Resources such as the NHGRI-EBI GWAS Catalog and the PGS Catalog already provide structured association and score data (59), (61). Evidence mining could connect variants to expression and protein QTLs, pathways, phenotypes, and clinical outcomes while preserving tissue, population, ancestry, allele, effect-size, and uncertainty information (60), (63)—important because predictive performance and transferability differ across populations (63).

Genetics also makes explicit a problem other domains let us postpone: much quantitative evidence lives not in running text but in long tables. A single study may discuss ten loci in text yet report tens of thousands of rows in supplementary spreadsheets, with column headers, allele conventions, scales, units, and thresholds that vary between studies. Reading such files row by row with a language model is neither feasible nor necessary. We propose a two-step approach: a model reads a small sample of rows to infer the schema (which columns contain the variant identifier, effect and non-effect alleles, effect estimate and scale, standard error, p-value, allele frequency, sample size, and genome build) and then applies that interpretation deterministically to the whole file. The model reads the schema, not the data.

Sampling also permits cheap plausibility checks: allele frequencies should be between 0 and 1, and when beta and standard error are reported, the test statistic ($z = \beta/SE$) and p-value should agree; odds ratios must be positive; and effect directions should be checked only after reconciling the effect alleles and genome build. Discrepancies are highlighted for manual inspection rather than quietly corrected.

## 8. Evaluation strategy and open benchmarks

Quantitative evidence mining should be evaluated at several levels. A system may detect a value correctly yet attach it to the wrong biomarker, unit, endpoint, population, or source; conversely it may extract a correct unit but miss that the value is implausible or inconsistent. A plausibility-aware system should therefore be evaluated across the levels in Table 5.

Table 5. Evaluation Levels and Example Metrics for Quantitative Evidence Mining

| Evaluation level | Metric examples |
|---|---|
| Quantity span extraction | Exact-match and overlap precision, recall, and F1 score |
| Unit extraction | Unit accuracy, unit-span F1 score, and unit-type accuracy |
| Entity/property linking | Relation precision, recall, F1 score, and attachment accuracy |
| Claim reconstruction and association | Subject–predicate–object F1 score, effect-direction accuracy, and claim–evidence association accuracy |
| Context extraction | Field-level F1 score, context-attachment accuracy, and missing-field detection accuracy |
| Normalization | Canonical-unit accuracy, numerical-conversion error, and scale accuracy |
| Provenance recovery | Source sentence, table, figure, and evidence-span alignment accuracy |
| Multidimensional assessment | Per-dimension macro-F1 score, calibration, area under the receiver operating characteristic curve (AUROC), expert agreement, and false-flag rate for valid outliers |
| End-to-end evidence-unit quality | Complete-unit accuracy, field completeness, expert acceptability, correction rate, and expert-review time |

Assessment performance should be reported separately for source grounding, contextual completeness, statistical coherence, biological plausibility, and cross-source consistency rather than summarized as one overall plausibility score.

Evaluation should distinguish between triple-level quality (subject, predicate, object, directionality, and source grounding) and evidence-unit-level quality (value, unit, measured entity and property, comparator, uncertainty, provenance, and context correctly extracted and linked). A system may correctly identify that "SARS-CoV-2 infection increases neuroinflammation" while failing to preserve the measured cytokine concentration, subgroup, time point, or uncertainty—so end-to-end evaluation should report both.

Existing benchmarks such as MeasEval provide an important starting point because they evaluate quantity spans, units, measured entities, measured properties, qualifiers, and relations between measurements and their context. However, biomedical quantitative evidence mining requires additional benchmark layers, including biomedical entity normalization, statistical-effect interpretation, source grounding, table and figure extraction, cross-sentence evidence linking, and expert assessment of plausibility. Future benchmarks should therefore evaluate complete evidence units rather than isolated quantity spans, and should include both text-based and multimodal evidence sources.

## 9. Discussion and outlook: beyond biomedical evidence

While this perspective focuses on biomedical evidence, the underlying phenomenon is broader: many domains are turning to AI to extract claims from long, complex, and partly unstructured documents. In each case, the interpretation of a claim often depends on numbers, units, thresholds, conditions, uncertainty, and provenance. The same consideration therefore applies beyond biomedical text mining: extracted claims should be measurable, traceable, contextualized, open to plausibility checks before being reused in downstream decisions.

### 9.1 Strengths and limitations of this review

A strength of this review is its integration of relation extraction, measurement extraction, multimodal document analysis, provenance, statistical coherence, biological plausibility, and downstream modelling within one auditable representation. The evidence-unit schema and layered architecture make the proposal concrete while keeping extraction confidence, contextual completeness, plausibility dimensions, and expert-review status separate.

Quantitative evidence mining should support, not replace scientific judgment. Errors in value extraction, unit conversion, contextual linking, and statistical interpretation can substantially alter a claim's meaning in clinical and regulatory settings. Evidence units should preserve their source links and uncertainty and remain open to expert inspection and correction.

In genetic applications, population-level associations should not be presented as individual risk estimates without appropriate validation, specialist interpretation, and governance. More generally, evidence units should not be treated as autonomous assessments of scientific truth. Their purpose is to make extracted claims easier to compare, trace, and audit while keeping human judgment within the workflow.

### 9.2 Research roadmap

The proposed framework remains conceptual and has not yet been implemented or validated as a complete end-to-end system. Its operationalization will require benchmark corpora that preserve claims, values, units, measured entities and properties, comparators, populations, experimental or clinical conditions, temporal context, uncertainty, and provenance across narrative text, tables, figures, and supplementary materials. Reference ranges and plausibility rules may not be available for every domain, and evidence generated using different assays, populations, endpoints, or study designs may not be directly comparable.

Future work should proceed in stages. First, annotation guidelines and representative benchmark datasets should be developed for different quantitative-evidence types. Second, individual components should be evaluated for quantity detection, entity and property linking, normalization, context association, claim reconstruction, and provenance recovery. Third, multidimensional assessment methods should be evaluated for calibration, sensitivity to genuine errors, and the risk of rejecting valid outliers. Finally, the complete workflow should be evaluated prospectively in clinical-trial analysis, biomarker synthesis, evidence-aware knowledge-graph construction, simulation-parameter retrieval, and quantitative genetic evidence mining.

## 10. Conclusions

Quantitative evidence mining extends biomedical text mining from relation extraction toward structured, source-grounded evidence representation. By preserving what was measured, how much it changed, under which conditions, with what uncertainty, and from which source, it makes machine-extracted claims more measurable, comparable, traceable, and auditable. The framework is a research agenda rather than a validated system; future work must operationalize and prospectively evaluate it with domain-specific benchmarks, multidimensional validation criteria, and expert-reviewed use cases.

## Acknowledgments

We thank Francesco Pappalardo and his team at the University of Catania, including Giulia Russo and Valentina Di Salvatore, for discussions on connecting quantitative evidence-based knowledge graphs with agent-based modelling. We also thank Kairntech for providing text-mining tool support within the broader collaboration underlying this work.

During manuscript preparation and revision, the authors used SciSpace and the research or deep-search functions available in ChatGPT (OpenAI; model/version not recorded) and Claude (Anthropic; model/version not recorded), accessed intermittently between July and September 2026, to support literature discovery and preliminary source summarization. ChatGPT and Claude were also used for language editing, structural refinement, and condensation of author-written text. Grammarly and QuillBot were used for grammar, spelling, style checking, and limited paraphrasing. Representative instructions included requests to identify potentially relevant literature, summarize specified publications, improve academic English, shorten passages without removing technical information, and reorganize author-written text for clarity. All suggested references, citations, bibliographic information, scientific claims, reported data, and quantitative values were manually checked against the original sources. AI-generated summaries were not treated as primary evidence. All outputs were reviewed and revised by the authors, who take full responsibility for the manuscript.

## Footnote

Reporting Checklist: The authors have completed the Narrative Review reporting checklist.

## Funding

This work was partly supported by the COMMUTE project, funded by the European Union under grant agreement 101136957. The project provided the collaborative context for the COVID-19–neurodegenerative disease comorbidity research discussed in this narrative review. The funder had no involvement in the conceptualization or preparation of the manuscript or in the decision to submit it for publication.

## Authors' Contributions

Conceptualization: NSB, SG, MCS, MHA, and MJ. Methodology: NSB, SG, MCS, MHA, and MJ. Resources: SG. Visualization: NSB. Writing – original draft: NSB. Writing – review and editing: NSB, SG, MCS, MHA, and MJ. Supervision: MHA and MJ. Project administration: NSB and MJ. Funding acquisition: MHA and MJ.

## Conflicts of Interest

SG is affiliated with Kairntech SAS, which provided text-mining tool support for the collaboration underlying this work. The remaining authors declare no conflicts of interest.

## Ethical Statement

Not applicable. This narrative review did not involve human participants, identifiable human data, or animals.

## Data Availability

No original datasets were generated or analyzed during the preparation of this narrative review.

## Abbreviations

AI: artificial intelligence

IL-6: interleukin-6

KG: knowledge graph

LLM: large language model

## Supplementary Appendix 1

Detailed examples and contextual facets of biomedical quantitative evidence units, including non-exclusive evidence facets and a structured BNT162b2 efficacy example. This material is provided as a separate Supplementary Appendix file uploaded with the submission.

### Supplementary Appendix 2

Complete PubMed search strategies, including database and platform details, execution date, coverage period, full queries, PubMed translations, warnings, limits and filters, retrieval and deduplication counts, and the seed-paper sensitivity check. This material is provided as a separate Supplementary Appendix file uploaded with the submission.